%% file: main.tex
\documentclass[preprint]{acmart}

\AtBeginDocument{%
  \providecommand\BibTeX{{%
    \normalfont B\kern-0.5em{\scshape i\kern-0.25em b}\kern-0.8em\TeX}}}

\setcopyright{none}
\usepackage{amsmath,amssymb,amsfonts}
\usepackage{algorithmic}
\usepackage{graphicx}
\usepackage{textcomp}
\usepackage{comment}
\usepackage{flushend}
\usepackage{subcaption}
\usepackage{gnuplottex}
\usepackage{xcolor}
\def\BibTeX{{\rm B\kern-.05em{\sc i\kern-.025em b}\kern-.08em
    T\kern-.1667em\lower.7ex\hbox{E}\kern-.125emX}}

\begin{document}

\newcommand{\eg}{\textit{e}.\textit{g}., }

\title{Exploiting Weight Redundancy in CNNs: Beyond Pruning and Quantization}

\author{Yuan Wen}
\email{weny@tcd.ie}
\affiliation{%
  \institution{Trinity College Dublin}
}

\author{David Gregg}
\email{david.gregg@cs.tcd.ie}
\affiliation{%
  \institution{Trinity College Dublin}
}

%%%\renewcommand{\shortauthors}{Trovato and Tobin, et al.}


\input{abstract}




\maketitle

\input{introduction}

\input{related_work}

\input{motivation}

\input{kernel_compression}

\input{experiment}

\input{huffman}

\input{conclusion}

\input{main.bbl}

\bibliographystyle{ACM-Reference-Format}
\bibliography{ref}

\end{document}

%% file: abstract.tex
\begin{abstract}
  Pruning and quantization are proven methods for improving
  the performance and storage efficiency of convolutional neural
  networks (CNNs). Pruning removes near-zero weights in
  tensors and masks weak connections between neurons in neighbouring
  layers. Quantization reduces the precision of weights by replacing
  them with numerically similar values that require less storage.
  In this paper we identify another form of redundancy in CNN weight tensors,
  in the form of repeated patterns of similar values. We observe that pruning
  and quantization both tend to drastically increase the number of repeated
  patterns in the weight tensors.

  We investigate several compression schemes to take advantage of this
  structure in CNN weight data, including multiple forms of Huffman coding,
  and other approaches inspired by block sparse matrix formats.
  We evaluate our approach on several well-known CNNs and find that we can
  achieve compaction ratios of 1.4$\times$ to 3.1$\times$ \emph{in addition} to
  the saving from pruning and quantization.

\end{abstract}

%% file: introduction.tex
\section{Introduction}
Deep Neural Networks are hugely successful in artificial intelligence
applications such as computer vision, natural language processing, and
robotics. Deep networks with a large number of layers and many
thousands of trainable parameters within in each layer can achieve
remarkable inference accuracy. However, these networks require large amounts of computation, memory and
energy~\cite{DBLP:conf/nips/HanPTD15} for inference. These heavy
requirements are a major barrier to the deployment of deep learning,
especially on resource-constrained mobile or embedded systems.

Although a large number of parameters can help to bring greater classification
accuracy, researchers have found that, in practice, parameters have a great
deal of redundancy.

In particular, many trained weights are close to zero, and it is often possible
to \textit{prune} these small weights (by setting them to zero) resulting in a
\textit{sparse} weight matrix~\cite{DBLP:journals/corr/IandolaMAHDK16,
DBLP:conf/cvpr/MaoHPLL0D17, DBLP:conf/iccv/LuoWL17}.

Weights which are not close to zero can still be \textit{quantized} to a lower
precision to reduce storage requirements. Quantization works by representing
fewer digits of, or eliminating, the fractional part of each weight. A number
of schemes have been proposed for both encoding and quantization of CNN
weights~\cite{DBLP:journals/cal/JuddAM17, DBLP:journals/corr/GyselMG16, 37631}.

Pruning and quantization can be enormously successful in reducing the
number and precision of weight parameters in DNNs. Researchers have
found that pruning can reduce the number of weights by up to 90\%
\cite{DBLP:conf/nips/HanPTD15}.

Despite the success of pruning and quantization, there is still a need to
reduce the size of weight tensors in DNNs. High-performance and efficient
inference on \emph{edge} devices is becoming crucial~\cite{tx2blog} to make
possible applications where response time is critical (e.g. detection of
pedestrians or obstacles in an automotive context). However, edge devices are
heavily constrained, particularly in terms of available memory for storing
weight tensors.

By examining weight tensors from various DNNs, we observed a great deal of
redundancy in the non-zero weights. We found that after pruning and
quantization, similar patterns of weights arise again and again, in both the
convolutional and fully connected layers of CNNs. This redundancy in
convolutional layers is particularly important because pruning is much less
effective in convolutional than fully-connected
layers~\cite{DBLP:conf/cvpr/MaoHPLL0D17}.

A typical approach to exploiting structural redundancy in data to reduce
storage requirements is to use a compression scheme to store data in memory.
Element-wise Huffman coding has previously been used to compress CNN weight
data~\cite{DBLP:journals/corr/HanMD15}, but other compression approaches seem
equally promising, particularly since the redundancy we observed in
weight data appears at a range of granularities, from single elements to whole
blocks of repeated weight data.

\subsubsection*{Contributions}

We make the following contributions:

\begin{itemize}
\item We study the prevalence of repeated patterns in CNN weight tensors, and
  show that there is significant redundancy even after pruning and quantization.
\item We evaluate both element-wise and block Huffman coding for weight compression.
\item We propose and evaluate a novel model compaction scheme that exploits redundancy
  in weight tensors represented in a block sparse format.
\item We evaluate our scheme and find that we achieve reductions of
  1.4$\times$ to 3.1$\times$ \textbf{in addition} to the savings from
  pruning and quantization.
\end{itemize}

%% file: related_work.tex
\section{Related Work}

DNN inference is often most useful in
real-time~\cite{DBLP:journals/corr/HanMD15} or
resource-constrained~\cite{DBLP:conf/micro/DingLWLLZWQBYMZ17,
DBLP:conf/cvpr/YangCS17} contexts. However, the computational complexity and
exceptionally large number of parameters in deep neural networks presents
challenges around execution time, data movement, and memory capacity in these
contexts~\cite{DBLP:conf/nips/DenilSDRF13}. Pruning and quantization both aim
to reduce the number of parameters in deep networks. Since the complexity of
most network layers is a function of the number of parameters, a reduction in
computation (typically stated as the number of multiply-accumulate or MAC
operations) accompanies parameter reduction~\cite{DBLP:conf/isca/YuLPDDM17}.

\subsection{Pruning}

Researchers have found that not all parameters make an equal contribution to
the output of any one DNN layer. Similarly, some connections between layers
have little impact on the output of the overall network. Removing (pruning)
these unimportant connections can save significant storage and reduce execution
time and has been widely advocated as an efficient method to reduce the number
of parameters~\cite{DBLP:conf/acssc/YangCES17, DBLP:journals/jetc/AnwarHS17,
DBLP:conf/nips/GuoYC16, DBLP:conf/nips/LinRLZ17, DBLP:conf/iccv/HeZS17}.

Much work on pruning focuses on identifying which weights can be pruned with
least effect on the classification accuracy of the overall network. Various
metrics, such as second-order derivative~\cite{DBLP:conf/nips/HassibiS92,
DBLP:conf/nips/CunDS89}, Average Percentage of
Zeros~\cite{DBLP:journals/corr/HuPTT16}, absolute
values~\cite{DBLP:journals/corr/LiKDSG16, DBLP:conf/nips/HanPTD15}, and output
sensitivity~\cite{pruning_based_onsensitivity}, have been proposed to guide the
pruning process.

Pruning results in a \textit{sparse} weight matrix, which can be compacted by
storing only the non-zero values~\cite{DBLP:journals/corr/abs-1708-02937}.
Common sparse matrix representations~\cite{DBLP:books/daglib/0009092} include coordinate (COO) format, where
each non-zero value is stored with its row and column coordinate; and
compressed sparse row (CSR) where non-zero values from the same row are grouped
together, and only the column index is stored for each non-zero.

These fine-grain sparse matrix formats save space, but modern CPUs and GPUs
provide vector SIMD/SIMT instructions that are much better suited to operating
on \emph{dense} matrix formats. Using a fine-grained sparse format typically
reduces computational performance versus similarly sized dense
matrices~\cite{DBLP:conf/isca/YuLPDDM17}, making high-performance
implementation of DNN layers more difficult.

To overcome this problem, alternative sparse matrix representations have been
developed, where the smallest granularity is a small dense block of data rather
than a single matrix element. The most widely-used of these formats is block
sparse row (BSR)~\cite{DBLP:conf/sc/WilliamsOVSYD07}, which is similar to CSR
but contains small dense blocks rather than indidual non-zeros.

\subsection{Quantization}

One way to reduce the storage required for non-zero values that remain after
pruning is to use \emph{approximate} values. Quantization is typically used for
inference, since at this stage the weight values are \emph{frozen}, and do not
need to track updates in high precision, as they do during the network training
process.

Rather than storing each value in the full precision that is used for training,
such as 32-bit floating point, a smaller size such as
16-bits~\cite{DBLP:journals/cal/JuddAM17},
8-bits~\cite{DBLP:journals/corr/GyselMG16, 37631}, or
4-bits~\cite{DBLP:conf/vlsic/MoonsV16} can be used for inference.

To convert the full-precision trained weights to lower-precision values for
inference, some \textit{quantization} scheme is needed. Provided the
quantization is not too severe, the loss in inference accuracy is typically
small~\cite{DBLP:journals/pieee/SzeCYE17}, but the saving in space is large.
For example, quantizing from 32-bit floating point to 8-bit integer reduces the
size of non-zero values by a factor of four.

\subsection{Encoding}

To further reduce the memory requirements for weight data, various encoding
schemes can be used. For example, Han et al.~\cite{DBLP:journals/corr/HanMD15}
use Huffman coding to compress the weight data even further. Huffman coding
works by building a dictionary of values in the input data, and replacing
instances of each particular value with that value's label from the dictionary.
The most frequent values are assigned the shortest labels. Using this tactic,
we can represent elements in the weight matrix using labels whose size is
related to the number of unique values.

%% file: motivation.tex
\section{Redundant Patterns}

When we examine the weights of a trained CNN, we observe many similar patterns.
The training process seldom creates patterns that are \emph{identical} to the
last bit of precision in every weight. However, pruning and quantization both
reduce the number of unique weight values appearing in the tensors. Two
patterns that are very similar before pruning and quantization often become
identical afterwards. The result is a large number of repeated patterns in the
weight tensors.

\begin{figure}[h]
\includegraphics[width=0.85\linewidth]{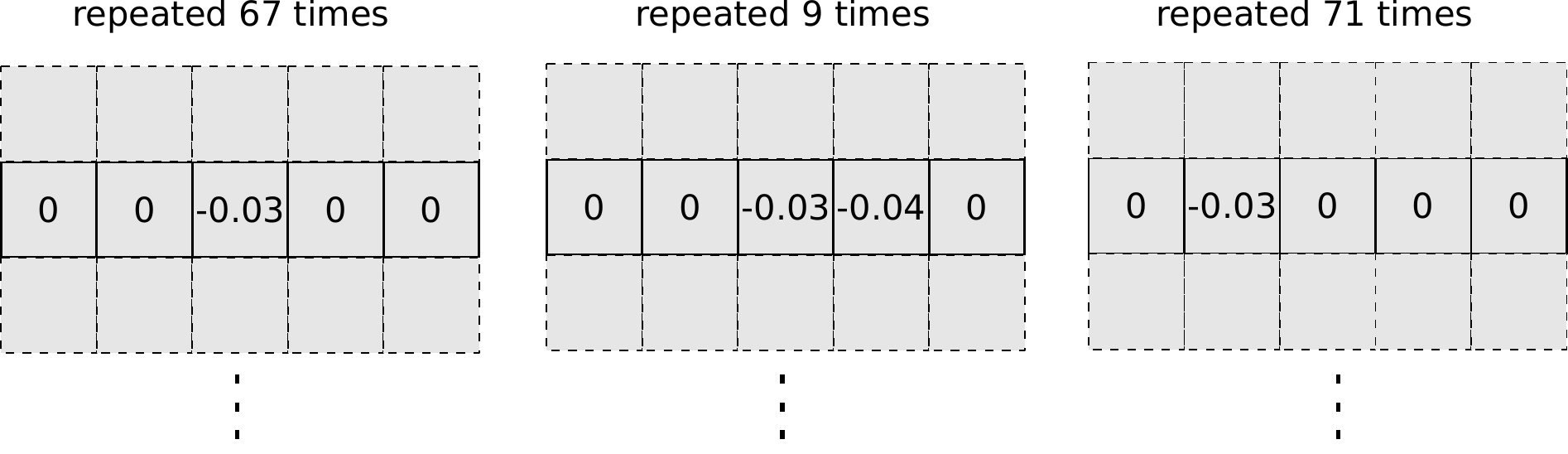}
  \caption{Repeated kernel-width vectors in the second convolutional layer of
  LeNet-5, after pruning and quantization}\label{fig:repeated_vectors}
\end{figure}

To illustrate this phenomenon, Figure~\ref{fig:repeated_vectors} shows
an example of the weight tensor of the second convolutional layer of
LeNet-5 after pruning and quantization.  In this convolutional layer
the kernel size of is $5 \times 5$, and for the purposes of
illustration we show the kernel tensor as a 2D matrix of width
5. Thus, each row of the matrix is one row of a $5 \times 5$
kernel. When viewed in this way, we can identify rows of the matrix
that appear more than once. Figure~\ref{fig:repeated_vectors} shows
three rows that appear 67, 9, and 71 times respectively. These
repeated rows offer opportunities for compacting the kernel tensor to
reduce memory requirements.

\begin{figure}[h]
  \includegraphics[width=0.55\linewidth]{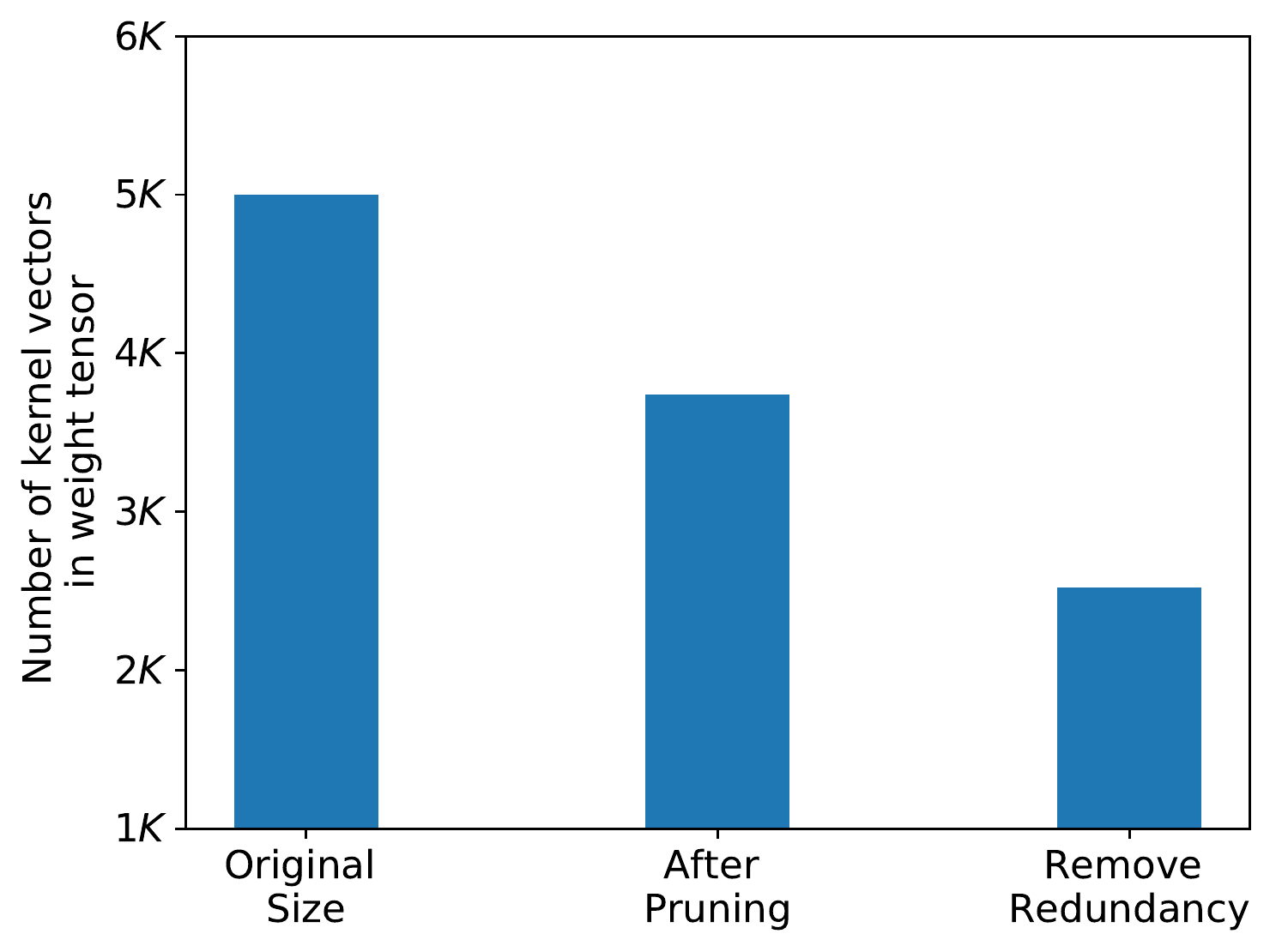}
  \caption{Parameter reduction in LeNet-5 by pruning followed by compression of redundant
  vectors.}\label{fig:motivation_statistic}
\end{figure}

In practice, the number of redundant vectors is large.
Figure~\ref{fig:motivation_statistic} shows the storage saving by keeping just
a single copy of each repeated vector. The first bar in
Figure~\ref{fig:motivation_statistic} shows the overall number of kernel-width
vectors in the second convolutional layer of LeNet-5. The second bar shows the
number of vectors after kernel-wise pruning and quantization. Finally, the
third bar shows the number of vectors remaining after removing repeated copies.
As we can see in this example, eliminating repeated patterns can provide an
additional saving of around 2$\times$ when compared with pruning and
quantization alone.

An important question is why so much redundancy arises between vectors of
trained weights. It can be difficult to fully understand why specific
parameters within a CNN receive a particular value during training. However, a
partial explanation is that CNNs learn to replicate aspects of classical
machine vision techniques.

\begin{figure}[t]
  \centering
  \includegraphics[width=0.7\textwidth]{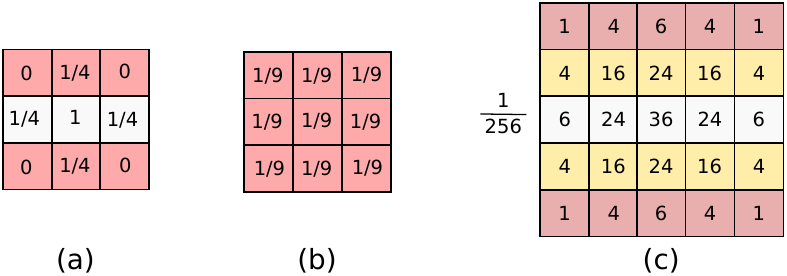}
  \caption{Kernels used for image processing~\cite{DBLP:books/lib/GonzalezW08}: (a) Edge detection (b) Box blur (c) Gaussian blur. As highlighted by rows, kernel-vector redundancy exists within kernels.}
  \label{fig:image_kernel}
\end{figure}

Figure~\ref{fig:image_kernel} shows three classical
machine vision filters that have been designed by humans to perform
edge detection and image blurring. It is notable that all three
kernels are symmetric along one or more axes. The symmetry of these
kernels introduces redundancies in a granularity of kernel-width. If
we remove repeated horizontal vectors from the kernels in
Figure~\ref{fig:image_kernel}, then the remaining values occupy
just two-thirds, one-third, and three-fifths of the orginal size
respectively.

The kernel values in a trained DNN are not designed by humans, but
instead emerge from the training process. The CNN learns them
iteratively by back propagation and stochastic gradient descent (SGD).
However, many of the same kernel features that are designed by humans
for classical machine vision are also likely to emerge from the
training process. These regular features are likely to appear
alongside other, more complex features that allow CNNs to exceed the
accuracy of classical vision techniques. Thus, we expect to see
symmetries emerge within trained kernels that can lead to repeated
rows within the kernel. Further, the same sub-features may appear
across multiple patterns leading to further redundancy. By keeping
just one copy of these common sub-features and sharing that copy among
multiple instances, significant space savings become possible.  In the
next section we explain how block sharing can reduce the size of CNN
models using a method inspired by \textit{block sparse row} (BSR)
format for representing sparse matrices.

%% file: kernel_compression.tex
\section{Model Compaction with Block Sharing}

\begin{figure}[h]
  \centering
  \includegraphics[width=0.8\textwidth]{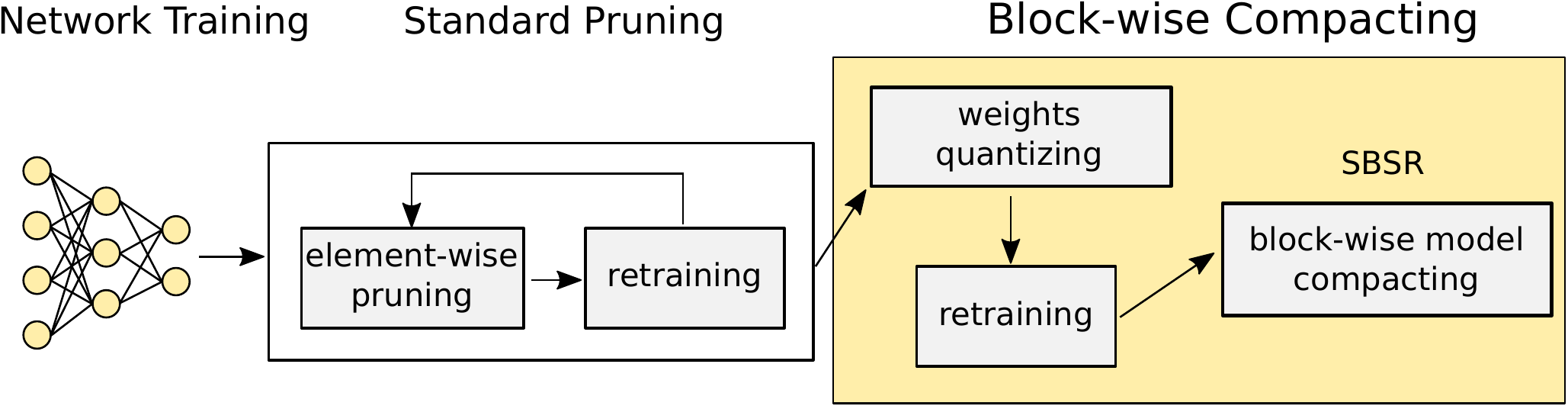}
  \caption{Flow of our DNN model compaction method}
  \label{fig:kernel_compression}
\end{figure}

Our block sharing method builds upon existing methods of network
pruning and quantization to further reduce the size of the model.  Our
method has four main steps, which are shown in
Figure~\ref{fig:kernel_compression}.

In the first step we prune the network to replace existing values with zero
where possible. We use the Scapel~\cite{DBLP:conf/isca/YuLPDDM17} pruning
method, which iteratively masks out values in the weight tensors and then
retrains the network to recover accuracy. The retraining step is critical to
the accuracy of the pruned model. We iteratively prune and retrain in a similar
way to other state-of-the-art~\cite{DBLP:journals/jetc/AnwarHS17,
  DBLP:journals/corr/LiKDSG16, DBLP:journals/corr/HuPTT16} pruning
techniques.

In the second step, we quantize the remaining non-zero weights to
reduce their precision, and thus the space required for storage.  Our
quantization factor is linked to the threshold value that is used in
pruning, so that losses in precision are similar for both
processes. After quantization, the network is again retrained to
recover the accuracy lost by reducing the precision of the weights.
Finally, we scan the weight tensors layer-by-layer to detect repeated
weight blocks and replace them with references to a single shared copy
of the block.

\subsection{Network Pruning}

Similar to the standard pruning methods, the network is first trained in full
32-bit floating point precision. The model is then iteratively pruned and
retrained. In the pruning step, a theshold value is selected and all weights
whose absolute value is below the threshold are tentatively masked to zero. The
network is then retrained to improve accuracy. In the forward step, masked
weights are treated as zeroes, but during back-propagation the original,
non-zero value is updated. Thus, a value that is pruned in one iteration may
recover in the next round of retraining.

One important question is the level of granularity at which pruning occurs. One
approach is to prune at the level of individual weights within a tensor.
Another method is to prune entire blocks of weights, or indeed entire kernels
or channels. In general, finer-grain pruning eliminates large numbers of
weights with little impact on the accuracy of the DNN, whereas a similar level
of coarser-grain pruning tends to have a large impact on accuracy
\cite{DBLP:conf/cvpr/MaoHPLL0D17}.  We prune at a fine grain to maintain
accuracy, but store the resulting tensors in a block-sparse row (BSR) format
which offers greater opportunity for efficient implementation on modern CPUs
and GPUs \cite{DBLP:conf/isca/YuLPDDM17}.

\subsection{Quantization and precision reduction}
The remaining non-zero weights are quantized and their precision is
reduced. An assumption of our pruning approach is that values smaller
than the threshold have only a minor impact on the result of the CNN
and can be safely removed. Similarly, in our quantization step we
may full-precision values to a nearby value that is representable
in lower precision. In our experiments we use 32-bit precision for
the original values, and 16-bit for the quantized values.

A question that is often ignored in discussions of quantization is the
rounding of full-precision values that fall between two representable
lower-precision values. The easiest strategy is to simply truncate
the lower bits of such values, but rounding to the nearest representable
value gives slightly better accuracy. The rounding strategy also has an
impact on the patterns of values that appear in the weight tensors,
and in the number of repeated patterns. We investigate this in more
detail in Section \ref{sec:experiment}.

Pruning and quantization are conceptually similar processes, in the
sense that they replace an exact value with a nearby approximation.
Just as we retrain after pruning, to maintain the accuracy of the CNN
we must also retrain after quantization, as shown in
Figure~\ref{fig:kernel_compression}. This process tunes the quantized
values and the bias to recover the accuracy of the model.

\subsection{Block sharing}
\begin{figure}[h]
\centering
\includegraphics[width=0.8\textwidth]{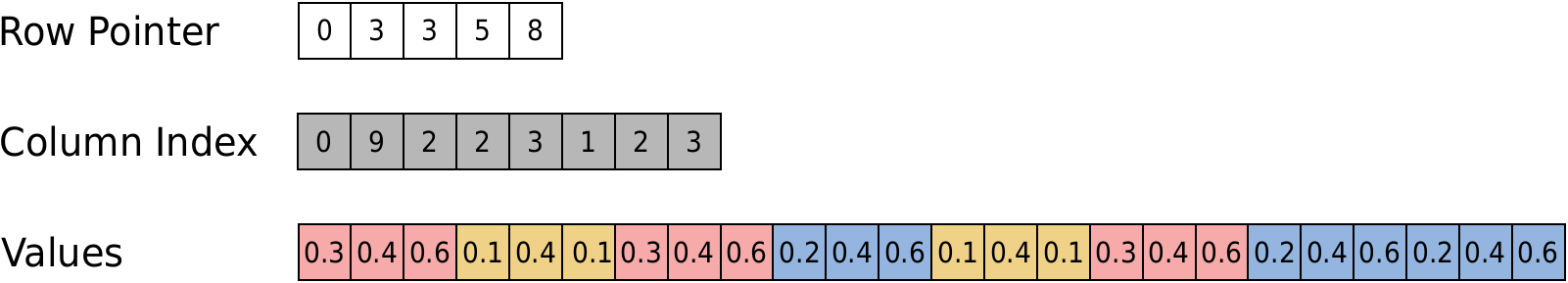}
\caption{Classic BSR format}
\label{fig:bsr}
\end{figure}

After pruning and quantization we represent the resulting matrix in
block sparse row (BSR) format (see Figure \ref{fig:bsr}).  In
contrast to fine-grain sparse formats, such as compressed sparse row
(CSR) format, BSR uses dense blocks of values containing at least one
non-zero rather than individual non-zeros. BSR has two main
advantages: it allows faster CPU SIMD and GPU implementations, and by
sharing the row and column coordinates between multiple separate
values it allows more compact matrix representations
\cite{DBLP:conf/cvpr/MaoHPLL0D17}.

\noindent Although BSR can result in more compact sparse matrix representations,
our results in this current paper show that it nonetheless contains a
great deal of redundancy. As we show in Section \ref{sec:experiment}
many instances of the same dense blocks occur many times in BSR
format.  We apply block-wise sharing to the matrix in BSR format to
eliminate this redundancy. We propose a new matrix format which we
call \textbf{shared-block sparse row} (SBSR) format, which allows repeated
blocks to be shared between different entries in the sparse matrix.

\begin{figure}[h]
\centering
\includegraphics[width=0.8\textwidth]{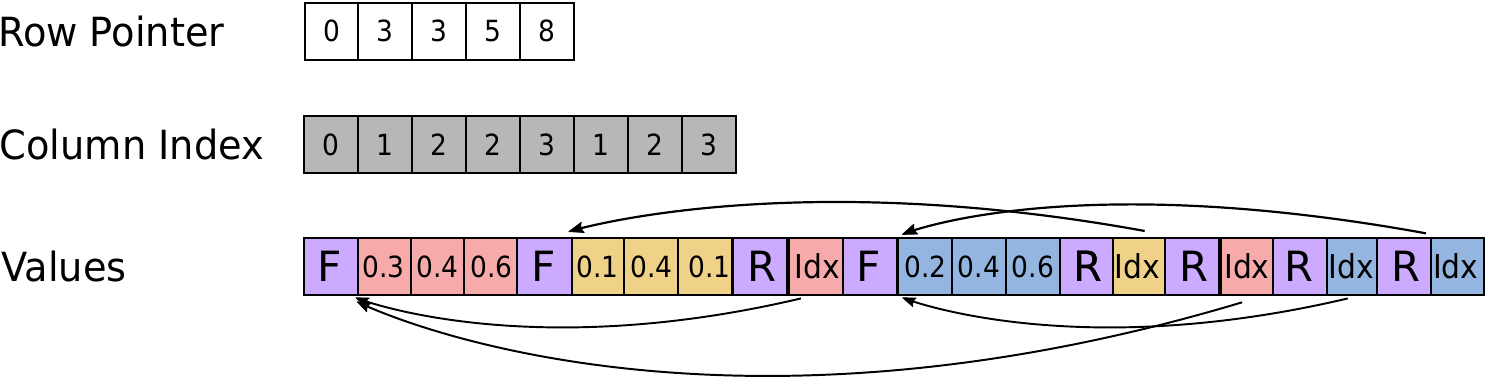}
\caption{SBSR format (ours)}
\label{fig:sbsr}
\end{figure}

\noindent Figure~\ref{fig:sbsr} shows our SBSR format. For blocks that appear
for the first time, the format is similar to BSR. The values of the
block are stored in a block vector, which exists alongside the row
pointers and column indices. However, when a block appears for a
second or subsequent time, the values of the block are not
represented. Instead, a reference is inserted into the block matrix,
which refers back to the previous location where that block appeared.
Thus, the values of a repeated block appear only in the first
appearance of that block, and subsequent appearances are replaced with
an reference to the shared block. Note that this format also requires
a flag to indicate whether the block appears for the first time (F in
Figure \ref{fig:sbsr}) or a repeat appearance (R). This flag can be
represented as a single bit.

%% file: experiment.tex
\section{Experimental evaluation}
\label{sec:experiment}

To evaluate our method we modified the Scalpel
\cite{DBLP:conf/isca/YuLPDDM17} framework for pruning and retraining
DNNs using an AMD Linux server with two Nvidia GTX 1080Ti GPUs. We set
the pruning thresholds to achieve target levels of sparsity, and added
a new quantization phase to reduce the precision of trained
weights. Finally, we build the resulting matrices in block sparse row
(BSR) and our own shared-block sparse row (SBSR) formats.

\begin{figure*}[t]
\centering
\includegraphics[width=0.9\textwidth]{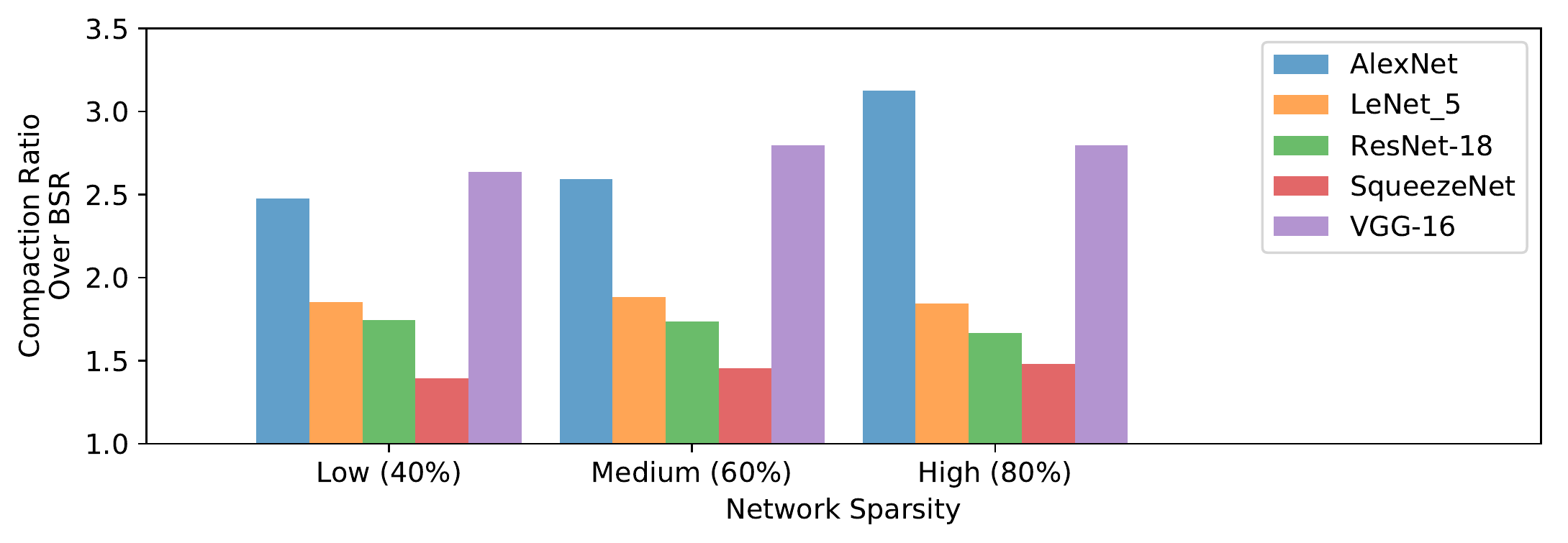}
\caption{Improvement of SBSR over BSR on CNNs for 40\%, 60\% and 80\% sparsity.}
\label{fig:result_summary}
\end{figure*}

Figure \ref{fig:result_summary} shows the factor reduction in size from sharing
repeated blocks rather than representing them each time they appear. We explore
three levels of sparsity: 40\%, 60\% and 80\%. Mao et
al.~\cite{DBLP:conf/cvpr/MaoHPLL0D17} found that pruning convolution kernels
beyond 40\%-60\% sparsity typically results in large losses in accuracy. In
contrast, fully connected layers can commonly be pruned to 80\%-90\% with
negligible loss of accuracy.

Figure \ref{fig:result_summary} shows that significant savings in
storage are possible using our shared-block strategy. For AlexNet, the
saving is a factor of around 2.4, 2.6 and 3.2 (which corresponds to a
reduction of around 58\%, 62\% and 69\%) in the size of the
represented matrix. The savings for other trained CNNs are smaller,
but in all cases the savings are positive and significant.
\begin{equation}\label{eq:size_bsr}
\small
Size_{BSR}={BSR_{idx}+BSR_{blocks}}
\end{equation}
\begin{equation}\label{eq:size_sbsr}
\small
Size_{SBSR} = {S_{flag}+S_{block\_pointer}+S_{idx}+S_{unique\_blocks}}
\end{equation}
~
\\
\noindent The memory requirement of BSR and SBSR formats are calculated according to
Equations \ref{eq:size_bsr} and \ref{eq:size_sbsr} respectively, where
$B_{idx}$ and $S_{idx}$ represent the storage for the BSR format column and row
indices, $BSR_{blocks}$ the storage for the all non-zero tensor blocks,
$S_{flag}$ the memory for storing the flag that indicates if the following
block is repeated or not, and $S_{unique\_blocks}$ is the size of all the
unique blocks that are present in the sparse tensor.

\begin{table}[h]
  \centering
   \caption{Breakdown of the compaction ratio by layer for AlexNet\label{tab:alexlayercompress} for 60\% sparsity}
  \begin{tabular}{ccccc}
   \toprule
	  \textbf{Layer} & \textbf{Dense} &  \textbf{Sparse Matrix} & \textbf{Sparse Matrix} & \textbf{Compaction}\\
	   & \textbf{Matrix}   & \textbf{(After Compaction)} & \textbf{(over BSR)} & \textbf{Ratio} \\
    \hline
	  conv1	&	90.75kB & 	40.03kB& 	 38.19kB & 1.05x		\\
	conv2   &	1200kB  &	626.7kB&	 386.9kB & 1.62x 	\\
	conv3   &	2.53MB  &	1.38MB &	 0.80MB	 & 1.73x 	\\
	conv4   &	3.38MB  &	1.12MB &	 0.67MB	 & 1.67x 	\\
	conv5   &	2.25MB  &	0.68MB &	 0.41MB	 & 1.66x 	\\
	fc6     &	144.0MB &	83.60MB&	 28.69MB & 2.91x 	\\
	fc7     &	64.00MB &	36.98MB&	 16.12MB & 2.29x 	\\
	fc8     &	15.63MB &	8.51MB &	 4.20MB	 & 2.03x 	\\
    \bottomrule
  \end{tabular}
  \label{tab:alexnet}
\end{table}

Table \ref{tab:alexnet} shows a more detailed breakdown of the compaction that
is achieved in different layers of AlexNet. We see that the level of block
sharing in the first layer, which is an $11 \times 11$ convolution, is very
small. However, the subsequent convolution layers, which use much smaller
kernels, offer much great opportunity for sharing blocks. Note that for
convolution layers, we use a block size that corresponds to one row of a
convolution kernel (i.e. a vector of length 11 for an $11 \times 11$ kernel).
The savings from sharing in the fully-connected layers are even larger. Table
\ref{tab:vgg} shows the same data for VGG16. There is a correlation between the
size of the matrix, and thus the number of blocks, and the opportunities for
sharing identical blocks.

\begin{table}[h]
  \centering
   \caption{Breakdown of the compaction ratio by layer for VGG16\label{tab:vgglayercompress} for 60\% sparsity}
  \begin{tabular}{ccccc}
   \toprule
	  \textbf{Layer} & \textbf{Dense} & \textbf{Sparse} & \textbf{Sparse Matrix} & \textbf{Compaction Ratio}\\
	   & \textbf{Matrix}  & \textbf{Matrix} & \textbf{(After Compaction)} & \textbf{(over BSR)}\\
    \hline
	  conv1\_1   &	6.75kB	& 3.32kB	& 3.12kB 	& 1.06x	  \\
	conv1\_2   &	144.0kB	& 71.93kB	& 50.10kB	& 1.44x	  \\
	conv2\_1   &	288.0kB	& 146.7kB	& 95.57kB	& 1.53x	  \\
	conv2\_2   &	576.0kB	& 203.5kB	& 179.5kB	& 1.13x	  \\
	conv3\_1   &	1.13MB	& 0.57MB	& 0.34MB 	& 1.70x	  \\
	conv3\_2   &	2.25MB	& 1.17MB	& 0.68MB 	& 1.74x	  \\
	conv3\_3   &	2.25MB	& 1.18MB	& 0.68MB 	& 1.74x	  \\
	conv4\_1   &	4.50MB	& 2.43MB	& 1.37MB 	& 1.77x	  \\
	conv4\_2   &	9.00MB	& 4.78MB	& 2.69MB 	& 1.76x	  \\
	conv4\_3   &	9.00MB	& 4.48MB	& 2.53MB 	& 1.77x	  \\
	conv5\_1   &	9.00MB	& 4.75MB	& 2.64MB 	& 1.80x	  \\
	conv5\_2   &	9.00MB	& 4.68MB	& 2.58MB 	& 1.81x	  \\
	conv5\_3   &	9.00MB	& 4.51MB	& 1.51MB 	& 2.99x	  \\
	fc6	   &	392.0MB	& 225.7MB	& 72.51MB	& 3.11x	  \\
	fc7	   &	64.00MB	& 37.30MB	& 14.99MB	& 2.49x	  \\
	fc8	   &	15.63MB	& 8.73MB	& 4.65MB 	& 1.88x	  \\
    \bottomrule
  \end{tabular}
  \label{tab:vgg}
\end{table}

\begin{figure}
\centering
\begin{subfigure}{0.45\linewidth}
\centering
\includegraphics[width=\textwidth]{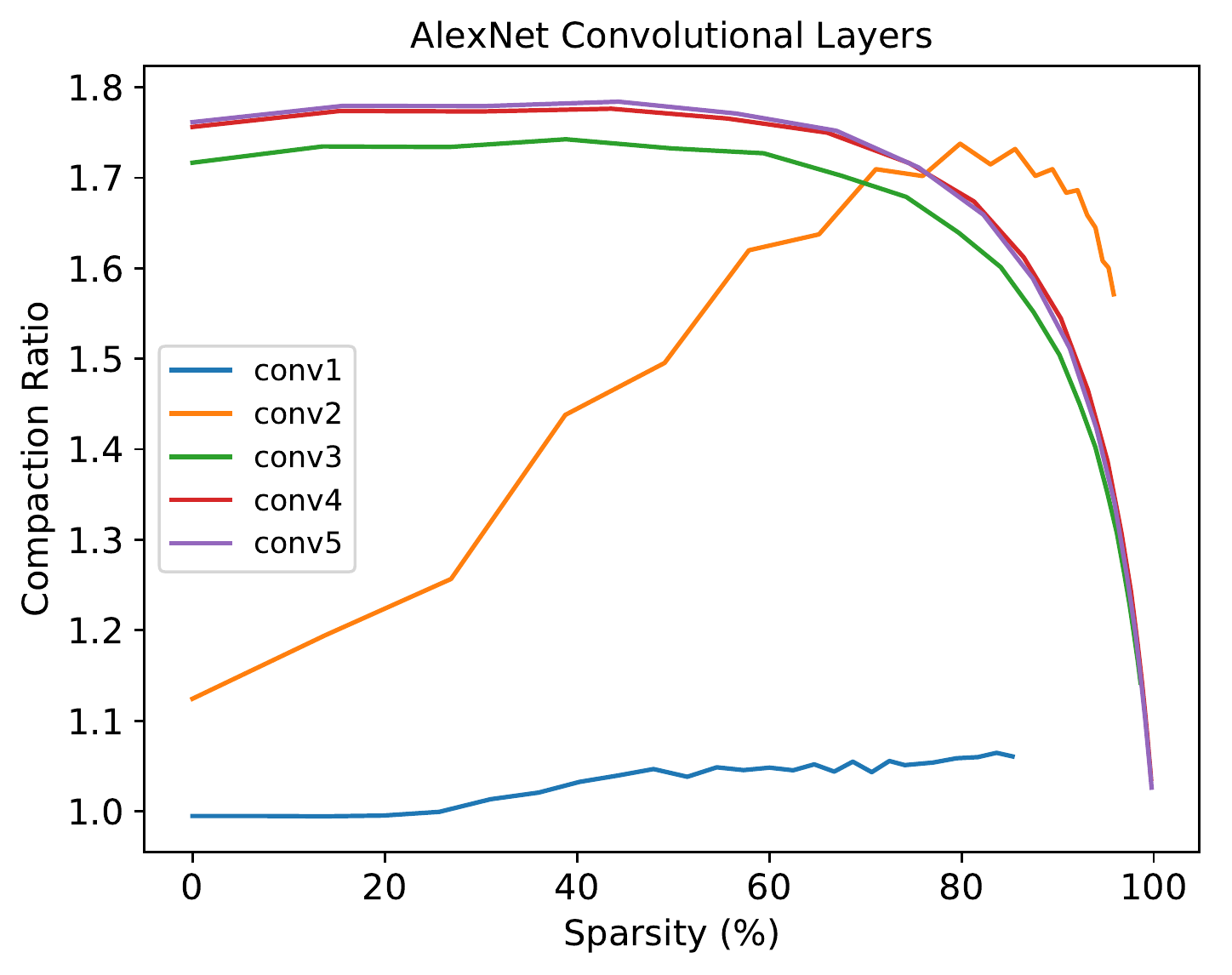}
\caption{Convolutional layers}
\label{fig:compressionAlexNetConv}
\end{subfigure}
\hspace{0.7cm}
\begin{subfigure}{0.45\linewidth}
\centering
\includegraphics[width=\textwidth]{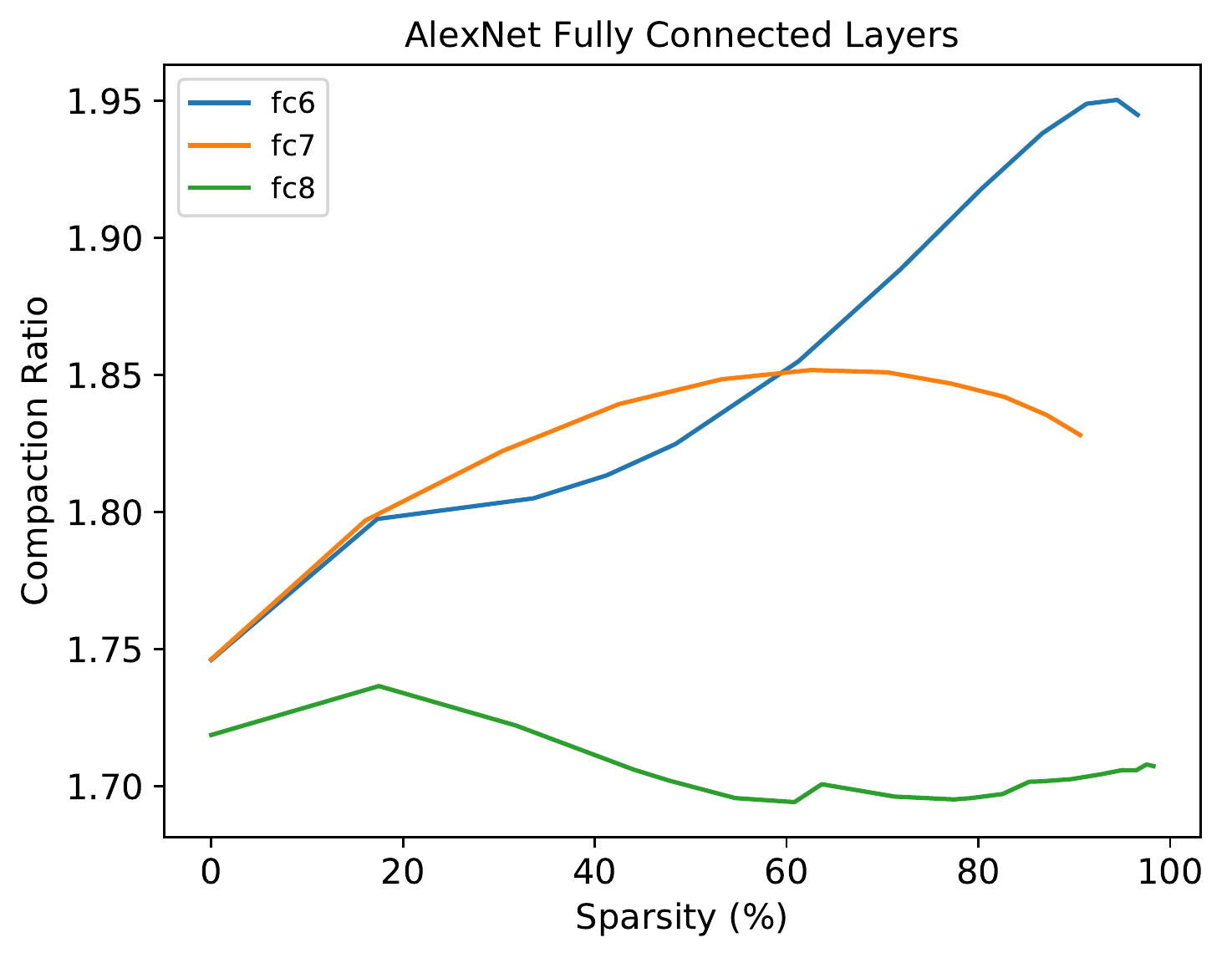}
\caption{Fully connected layers}
\label{fig:compressionAlexNetFC}
\end{subfigure}
\caption{Compaction per layer for AlexNet}
\label{fig:alexnet-graph}
\end{figure}

\begin{figure}
\centering
\begin{subfigure}{0.45\linewidth}
\centering
\includegraphics[width=\textwidth]{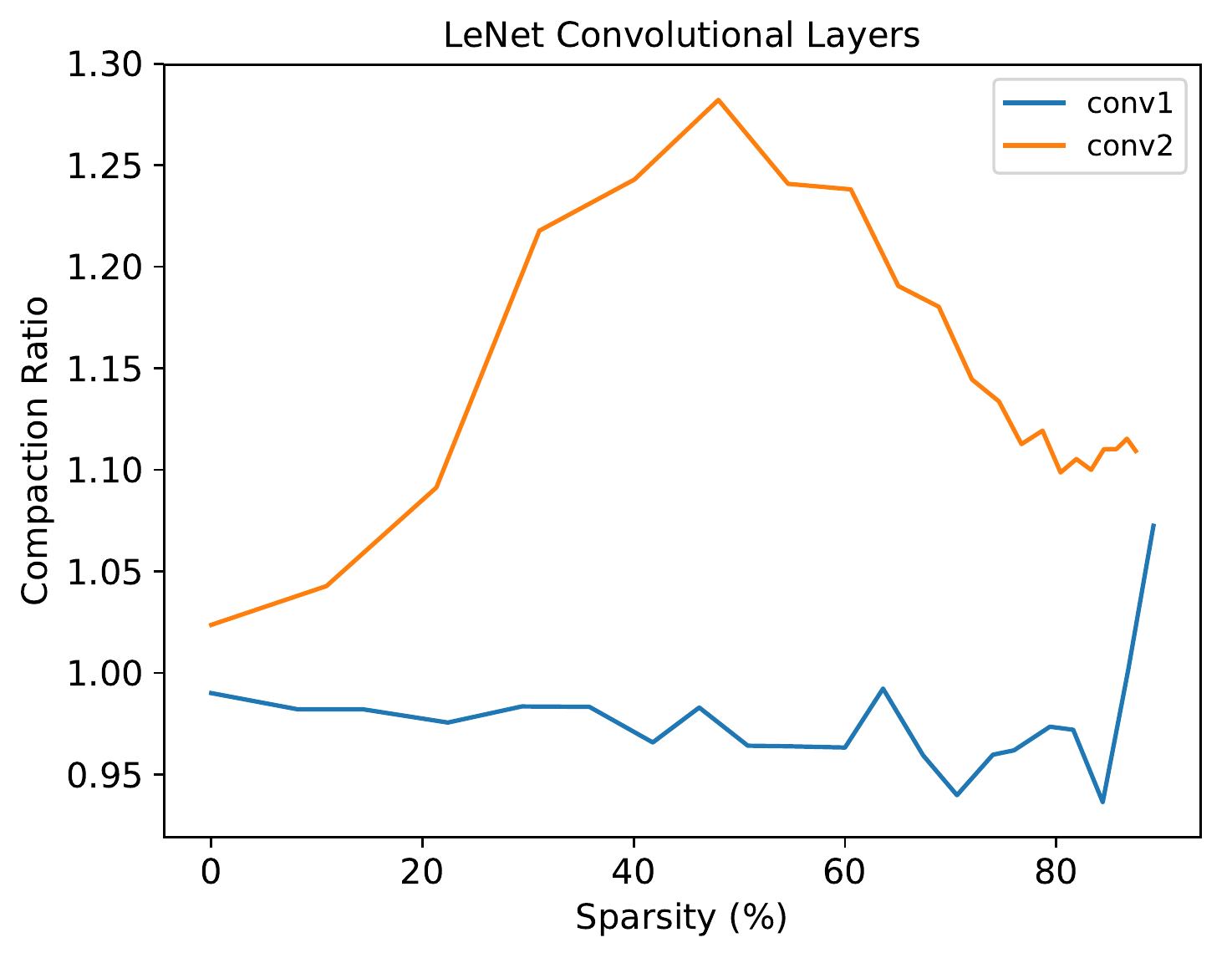}
\caption{Convolutional layers}
\label{fig:compressionLeNetConv}
\end{subfigure}
\hspace{0.7cm}
\begin{subfigure}{0.45\linewidth}
\centering
\includegraphics[width=\textwidth]{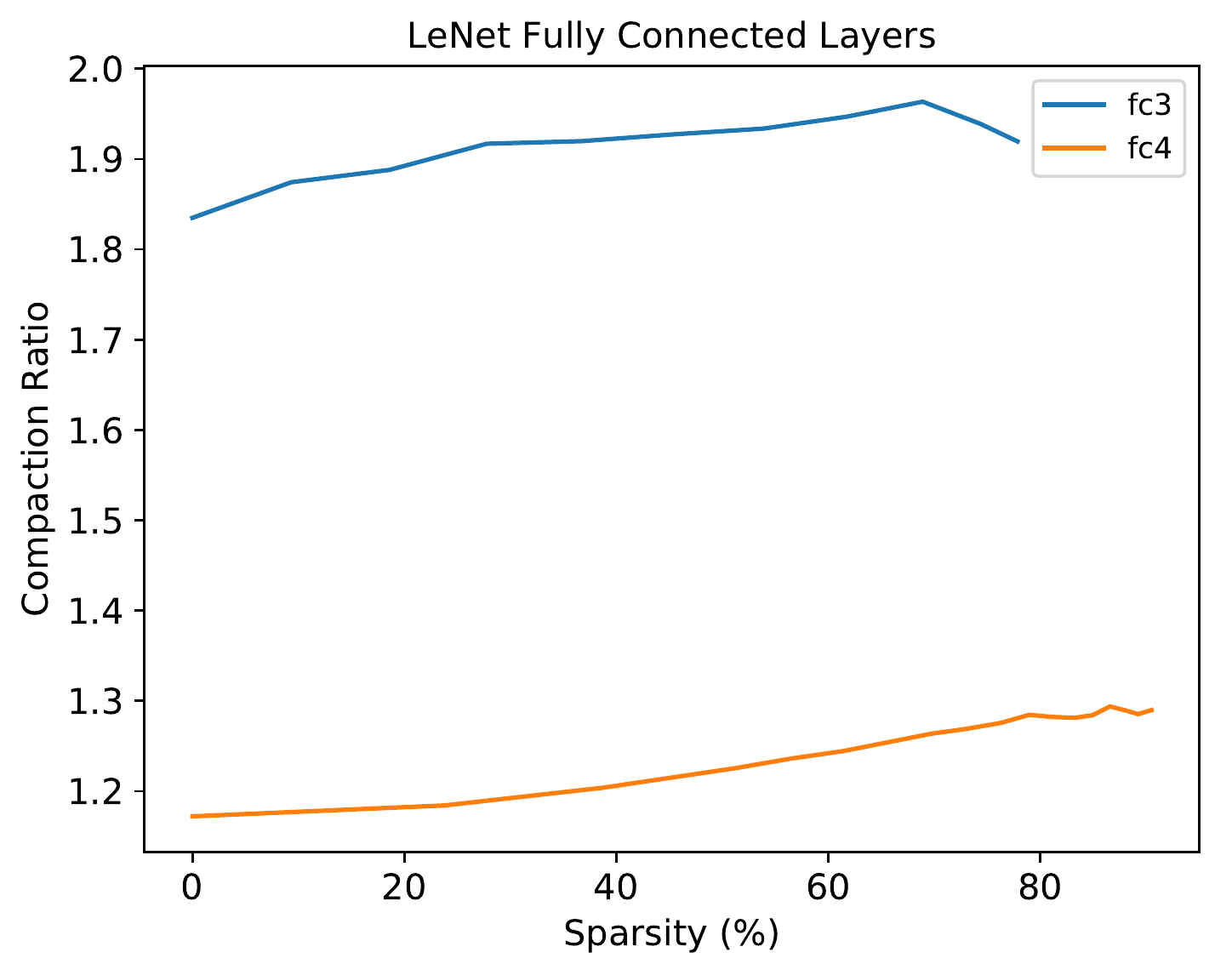}
\caption{Fully connected layers}
\label{fig:compressionLeNetFC}
\end{subfigure}
\caption{Compaction per layer for LeNet}
\label{fig:lenet-graph}
\end{figure}

Figure \ref{fig:alexnet-graph} shows another view of the compaction ratio for
different layers of AlexNet. Figure \ref{fig:compressionAlexNetConv} shows the
compaction ration for AlexNet's five convolution layers. The rows of the $11
\times 11$ kernels of the \texttt{conv1} layer provide few opportunities for
sharing. The rows are too long, the values are too diverse, and the number of
the kernels too small for many repeated rows to appear. In contrast, the level
of sharing increases rapidly as sparsity increases in the weights for layer
\texttt{conv2}. As more small values are replaced with zero, small differences
between blocks tend to disappear and more sharing becomes possible. In
contrast, convolution layers 4, 5, and 6 have a great number of repeated blocks
even without pruning. The compaction ratio for these layers \emph{falls} with
very high levels of sparsity simply because blocks that might otherwise be
duplicates are eliminated entirely when all values are replaced with zeroes.
The block sharing in the LeNet convolution layers
(Figure~\ref{fig:compressionLeNetConv}) follows a similar pattern to the first
two layers of AlexNet.

The AlexNet fully-connected (FC) layers (Figure~\ref{fig:compressionAlexNetFC})
exhibit high levels of block sharing, which is consistent with the large size
and large numbers of blocks in these layers. In LeNet, which has much smaller
FC weight tensors, the level of sharing is much less consistent.

For convolutional layers, we select a block size that is equal to the size of
one row of a kernel. This allows our method to benefit from repeated patterns
across different kernels. However, for fully-connected layers, the appropriate
block size is less clear.  A small block size tends to result in a great many
repeated blocks, which reduces the space needed to store the unique blocks.
However, each non-zero block needs a column index for its location, and
repeated blocks need an index that refers to the location of its shared block.
Thus very small blocks can be quite space inefficient. Using a larger vector
block size tends to result in less sharing of common blocks, but requires less
space for indices.

\begin{table}[h]
  \caption{Optimal vector size for FC layers at 60\% sparsity}
  \begin{tabular}{ccccc}
  \toprule
  \textbf{Network} & \textbf{Layer} & \textbf{Block} \\
  & & \textbf{Size} \\
  \hline
  AlexNet & fc6 & 8\\
  AlexNet & fc7 & 8\\
  AlexNet & fc8 & 4\\
  VGG16 & fc6 & 8\\
  VGG16 & fc7 & 4\\
  VGG16 & fc8 & 4\\
  ResNet & fc & 2\\
  LeNet & fc3 & 4\\
  LeNet & fc4 & 2\\
  \bottomrule
  \end{tabular}
  \label{tab:vectorsizefclayers}
\end{table}

\begin{figure}
\centering
\begin{subfigure}{0.45\linewidth}
\centering
\includegraphics[width=\textwidth]{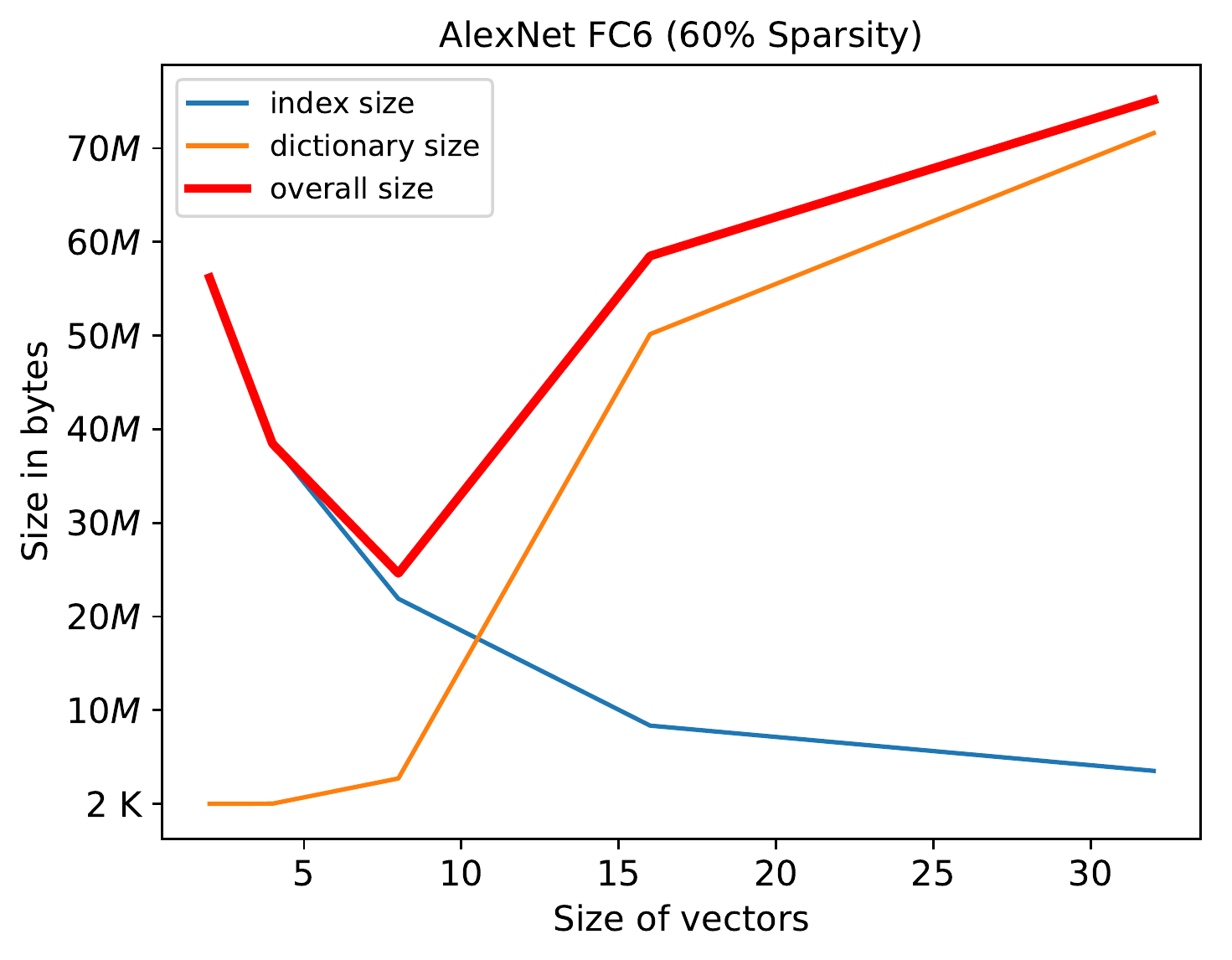}
\caption{AlexNet FC6}
\label{fig:costCompressionFC6}
\end{subfigure}
\hspace{0.7cm}
\begin{subfigure}{0.45\linewidth}
\centering
\includegraphics[width=\textwidth]{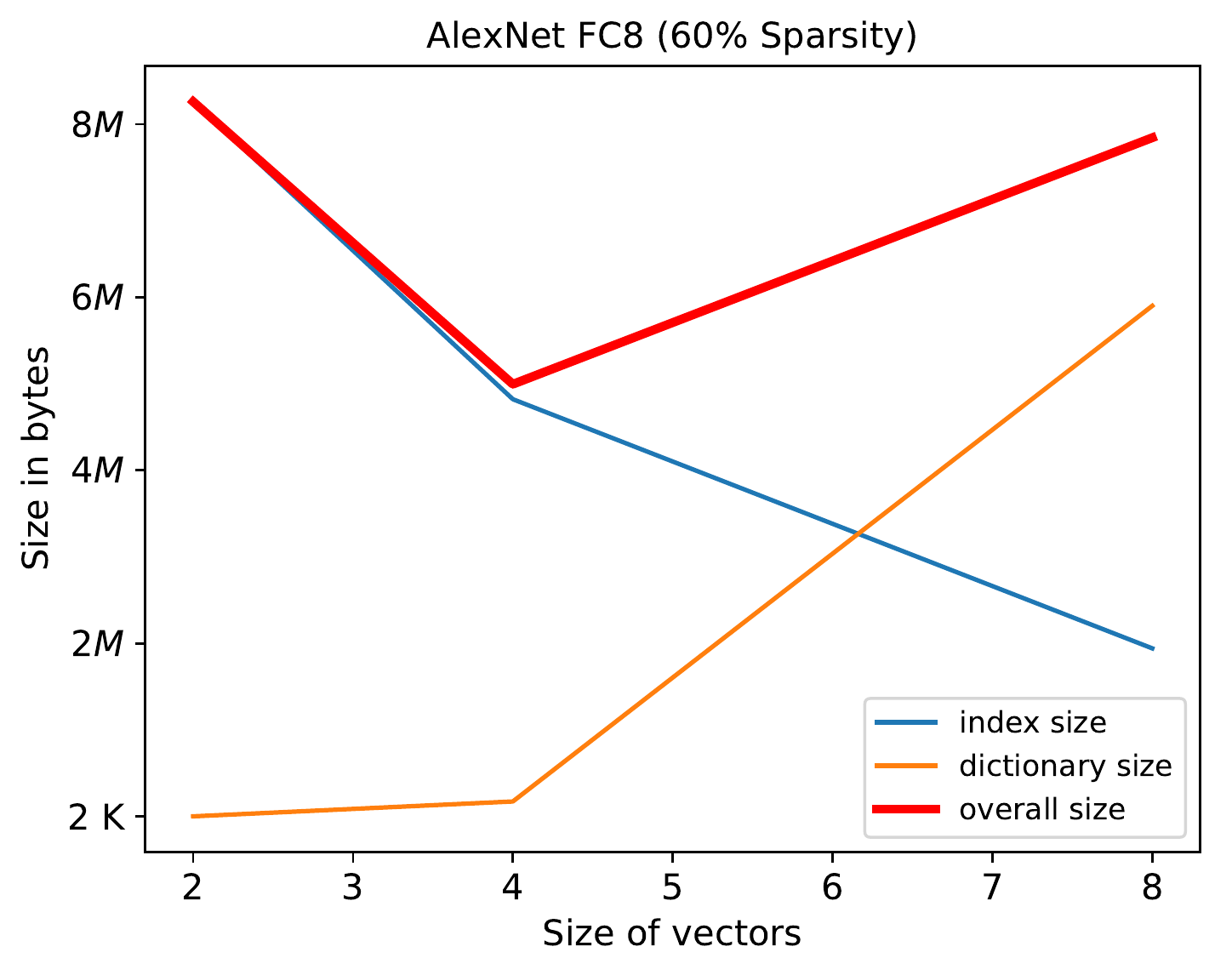}
\caption{AlexNet FC8}
\label{fig:costCompressionFC8}
\end{subfigure}
\caption{Impact of vector sizes on model compacting. The curves are the lower the better. Compaction ratio goes worse with the vector size increasing. However, the cost introduced by index decreasing. Table \ref{tab:vectorsizefclayers} shows the best recorded vector sizes.}
\label{fig:cost_for_compression}
\end{figure}

Figure \ref{fig:cost_for_compression} shows the trade-off between block storage
and index storage for AlexNet layer \texttt{fc6} and \texttt{fc8}. In both
cases the best block size is a compromise between block and index storage, with
a size of four for \texttt{fc6} and eight for \texttt{fc8}. Table
\ref{tab:vectorsizefclayers} shows the optimal block size for fully-connected
layers across several different CNNs. In general it seems that FC layers with
more parameters tend to benefit from larger block sizes.

\begin{figure}
\centering
\begin{subfigure}{0.45\linewidth}
\centering
\includegraphics[width=\textwidth]{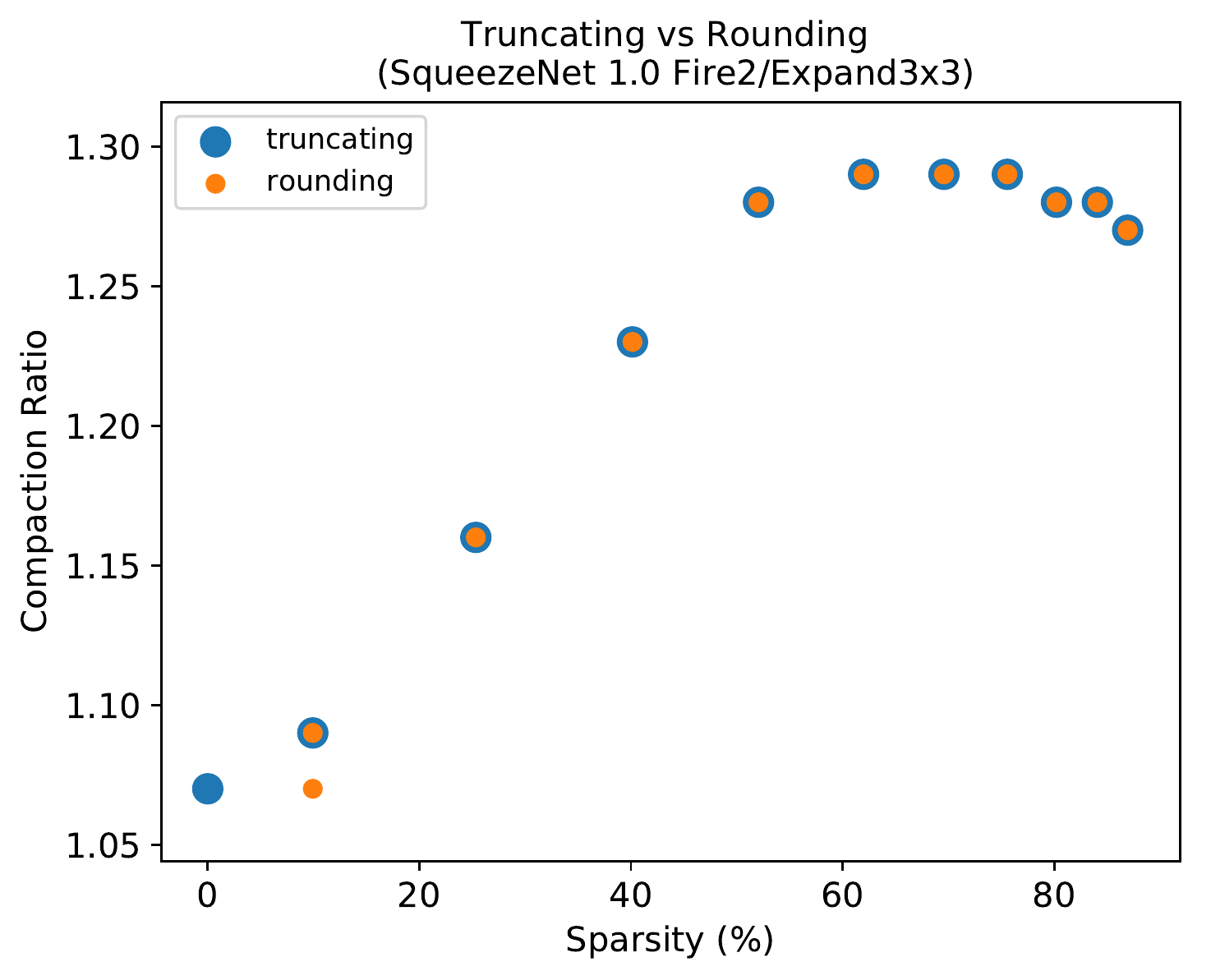}
\caption{'squeeze' layer from SqueezeNet 1.0}
\label{fig:truncationVsRoundingSqueezeNet}
\end{subfigure}
\hspace{0.7cm}
\begin{subfigure}{0.45\linewidth}
\centering
\includegraphics[width=\textwidth]{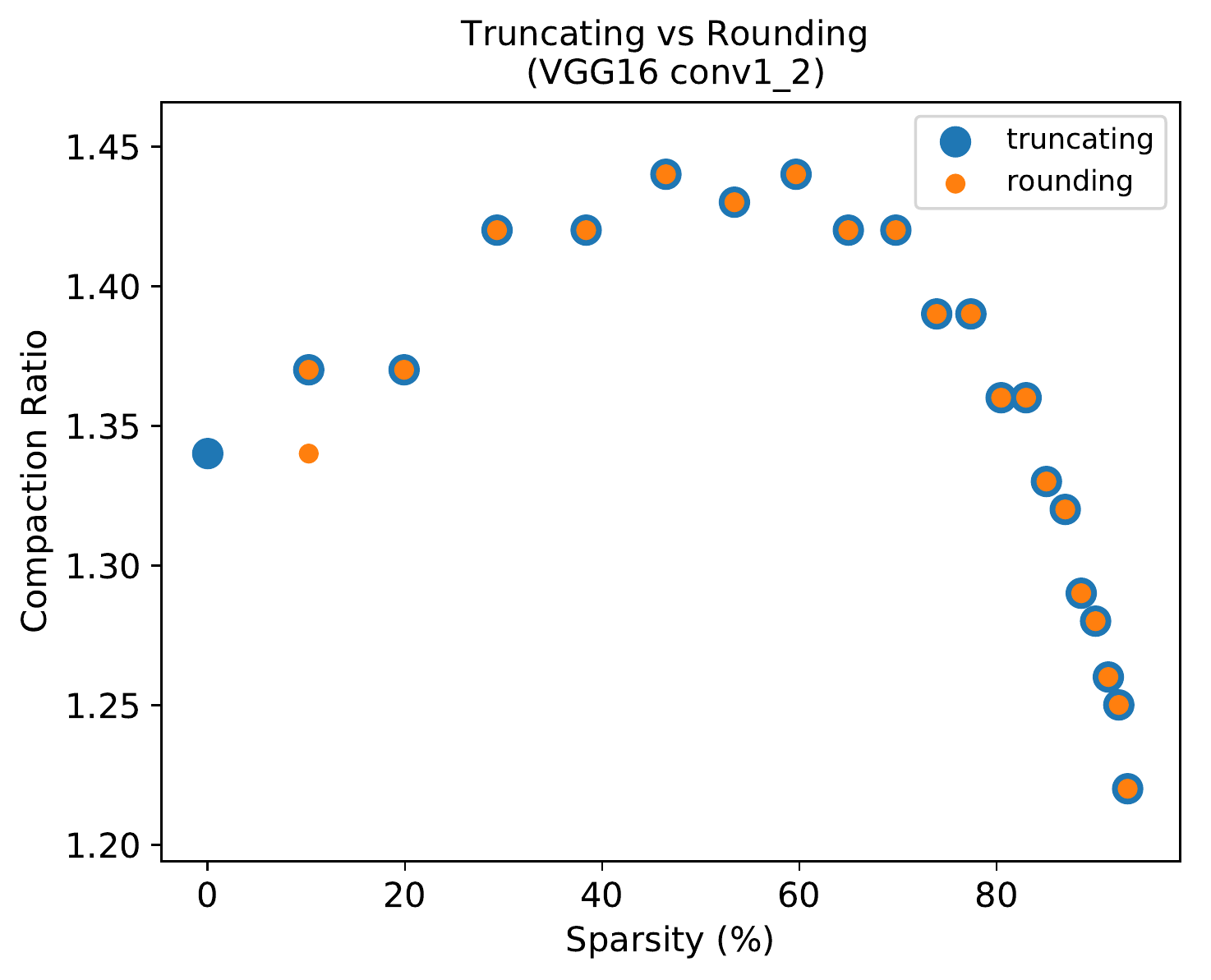}
\caption{Convolutional layer from VGG16}
\label{fig:truncationVsRoundingVGG}
\end{subfigure}
\caption{Comparison between the compaction ratio when using truncation and rounding}
\label{steady_state}
\end{figure}

Finally, we investigated the effect of either quantizing the weights
by rounding to the nearest representable value or by simple
truncation. The results show that both approaches provide almost
identical levels of block sharing. Given that rounding to the nearest
value gives a slightly higher accuracy, this is the method that should
be used.

%% file: huffman.tex
\section{Vector Sharing vs Element Sharing}

\begin{figure*}
\centering
\begin{subfigure}{0.95\linewidth}
\centering
\includegraphics[width=\textwidth]{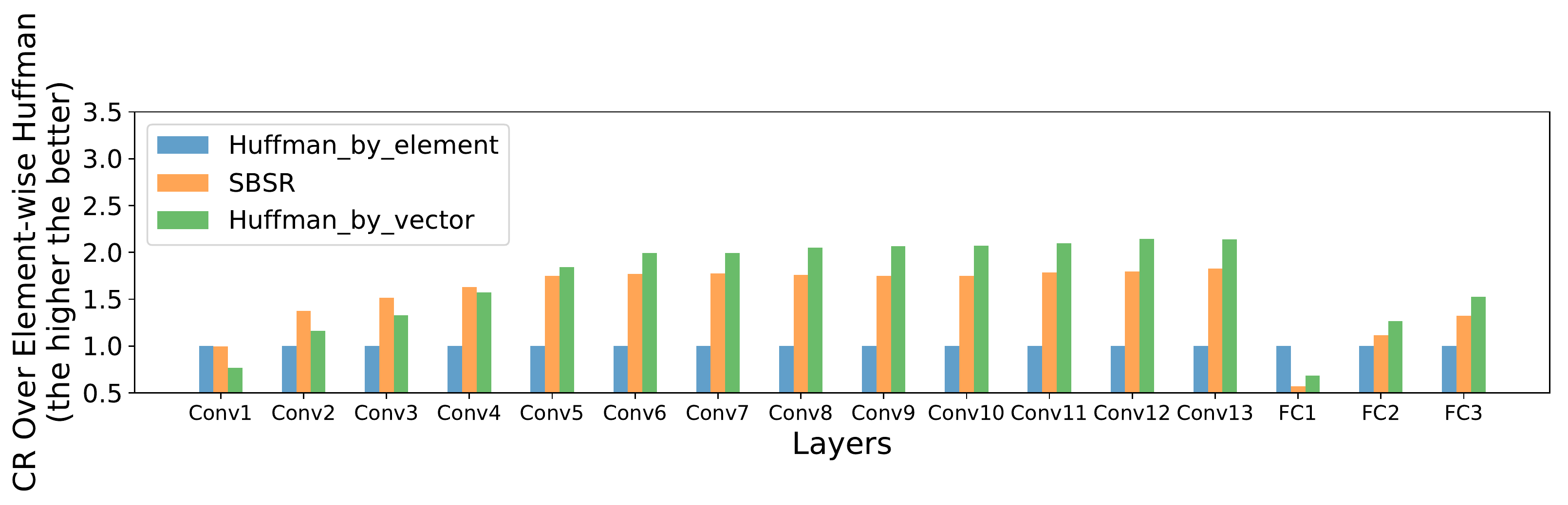}
\caption{Normalized size of memory consumption on VGG16}
\label{fig:huffman_vgg16_f3}
\end{subfigure}
\hspace{0.7cm}
\begin{subfigure}{0.95\linewidth}
\centering
\includegraphics[width=\textwidth]{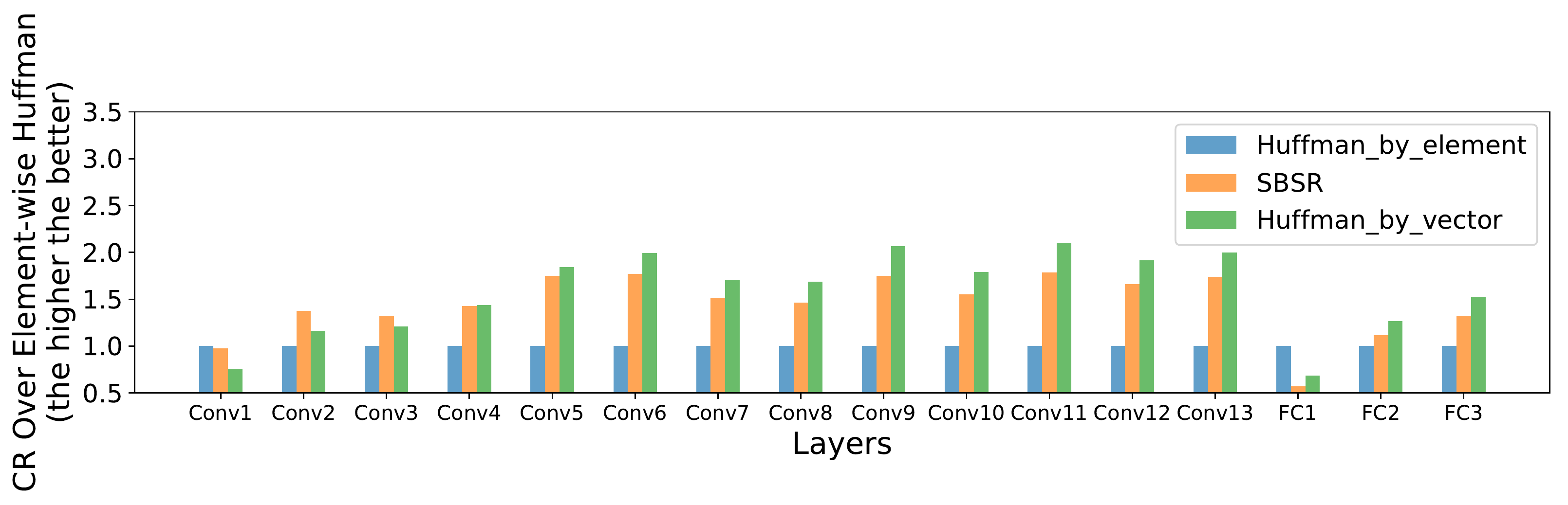}
\caption{Normalized size of memory consumption on VGG16, but 10-20\% Sparser than~\ref{fig:huffman_vgg16_f3}}
\label{fig:huffman_vgg16_f6}
\end{subfigure}
\caption{Comparison between element-wise and vector-wise Huffman coding on
 VGG-16}
\label{huffman_vgg16}
\end{figure*}

For compression of the weight data after pruning, Huffman coding is a
widely advocated method. However, existing work focuses on element-wise
encoding, e.g. Deep Compression~\cite{DBLP:journals/corr/HanMD15} which
replaces every weight with a variable-length code to reduce the size of the
sparse tensor. Though effective, element-wise Huffman coding works suboptimally
in many cases because it ignores repeated \emph{patterns} of values, leading to
missed opportunities for compression.

While SBSR works conceptually like a linked list of blocks, the Huffman coding
algorithm encodes symbols by building a binary tree according to their
occurrence frequencies. All symbols are leaves of the tree while the path to a
given leaf is its Huffman code. Rather than assigning every symbol a fixed
length code, Huffman coding introduces a variable-length code which allocates
fewer code bits to symbols that occur more often. The overall size of the
memory requirement is therefore reduced by the use of shorter codes for higher
frequency values.

It is straightforward to see how a block-sparse representation exploits
patterns in nonzeroes to increase the efficiency of storage. Each index incurs
some overhead, so when indices address blocks of data, rather than single
elements, the overhead is amortized by the size of the blocks, at the expense
of a slight increase in the number of stored zeroes in partial blocks.

In order to capture repeated patterns of values using Huffman coding, we
propose vector-wise Huffman coding, which assigns a code to a vector of values,
as opposed to individual matrix elements. To fully understand the advantage of
vector-wise coding over element-wise coding, we do a breakdown analysis. We
then compare the memory reduction from SBSR, vector-wise Huffman coding and
element-wise Huffman coding.

The storage required by element-wise Huffman coding
consists of three parts: indices, variable-length codes and the encoding dictionary. The
memory requirement is calculated as shown in Equation~\ref{equ:huffman}.

\begin{equation}
\small
Huff\_Size = H_{Idx}+H_{dict}
+\sum_{i=1}^{Size\_of\_Dict}{code\_length_{i} \times freq_{i}}
\label{equ:huffman}
\end{equation}

The indices $H_{Idx}$ represent the row and column index of the encoded data in
the weight matrix. For element-wise Huffman coding, the value indexed is a
single non-zero matrix element, while for vector-wise encoding, the value is a
weight \emph{vector}. $H_{Dict}$ is a lookup table with two columns, with one
column listing Huffman codes and the other presenting the original value. It
contains the necessary information for decoding the weights.

After replacing each element of the tensor with its Huffman code, the memory
required to store the encoded values is the sum of the code size times its
frequency. We use the formula $\sum_{i=1}^{Size\_of\_Dict}{code\_length \times
freq}$ to denote this.

To compare vector- and element-wise encoding methods, we use element-wise
Huffman coding~\cite{DBLP:journals/corr/HanMD15} as our baseline.
We use the Compaction Ratio (CR) on graphs as a metric to evaluate the comparison.
The CR is read as the ``improvement'' in compression versus element-wise
Huffman coding (so larger CR represents greater compression).

We calculate CR as shown in Equations \ref{equ:RNCR_huffman} and \ref{equ:RNCR_huffman_SBSR}.
\begin{equation}
\small
	CR\_huffman = \frac{Size\_of\_Element\_Wise\_Huffman}{Size\_of\_Vector\_Wise\_Huffman}
	\label{equ:RNCR_huffman}
\end{equation}

\begin{equation}
\small
	CR\_SBSR = \frac{Size\_of\_Element\_Wise\_Huffman}{Size\_of\_SBSR}
	\label{equ:RNCR_huffman_SBSR}
\end{equation}

Figure~\ref{huffman_vgg16} presents the result of vector-wise over element-wise compaction.
Limited by the length requirement of the paper, we only present the experiment
carried out on the VGG-16 network.
Similar results have been found across other networks in fact.
All of the 16 layers, including 13 convolutional layers (Conv) and 3 fully connected (FC) layers, have been examined in our experiment.
Besides all layers, the network under two different sparsities is also examined.
Though the sparsity varies across layers in practice, on average, each layer in Figure~\ref{fig:huffman_vgg16_f3} is 10-20\% sparser than in Figure~\ref{fig:huffman_vgg16_f6}.

\begin{figure}[t]
  \includegraphics[width=.55\linewidth]{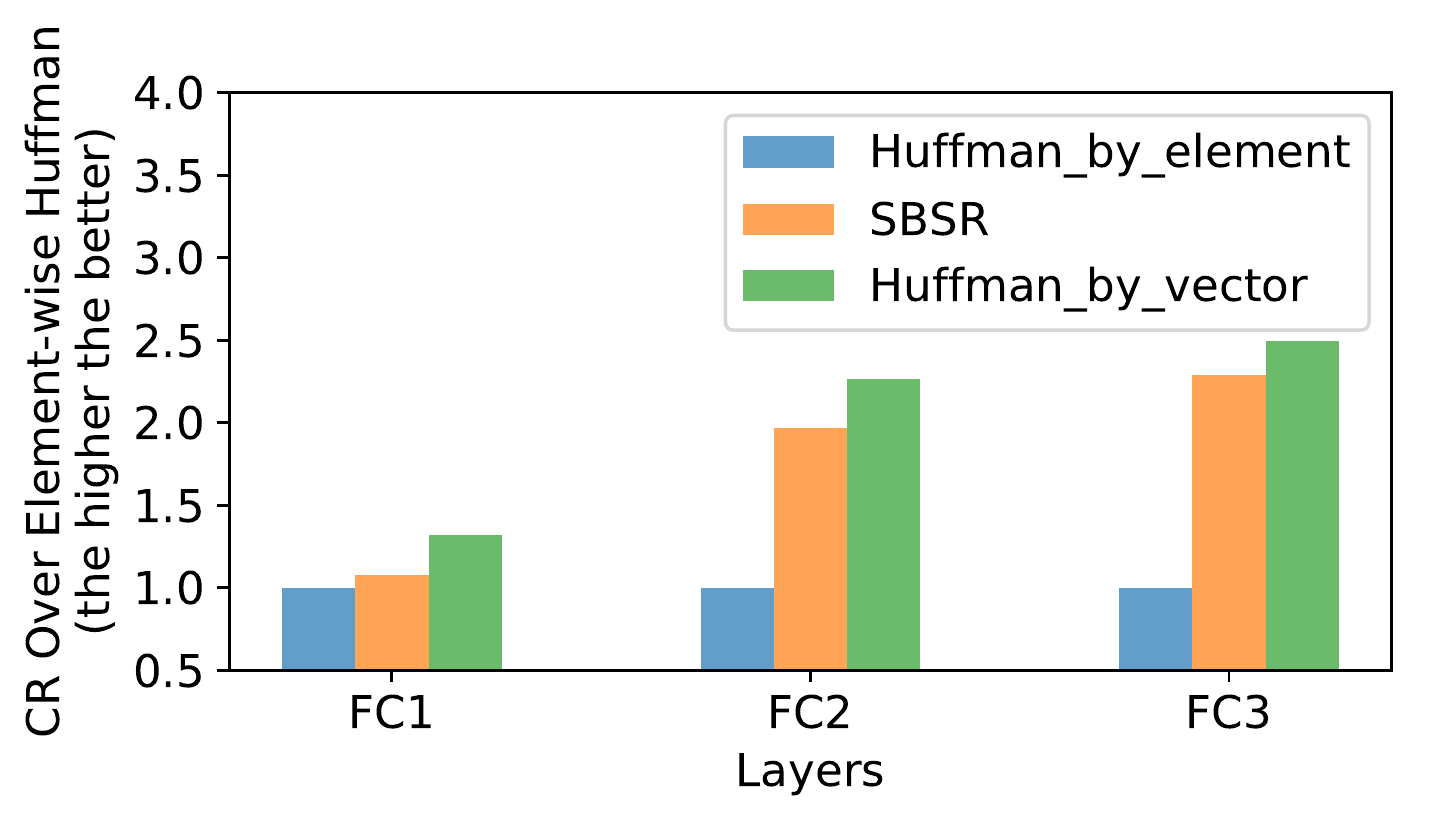}
  \caption{Using a larger vector to the fully connected layers. In this experiment, the size of vector is 4.}\label{fig:huffman_vgg16_coarse_vector}
\end{figure}

As presented in Figure~\ref{huffman_vgg16}, in most cases, the vector-wise sharing works better than element-sharing.
There are two exceptions, which are the layers Conv1 and FC1 in our experiment.
The front layers usually come with low sparsities.
In our experiment, the sparsity of the Conv1 is the lowest.
Because the length of the vector-based dictionary is much larger than element-based one for a dense matrix, extra code bits and storage space are required accordingly.
Therefore, the element-wise Huffman coding works better than the vector-wise implementation.
However, as SBSR does not require encoding, it works equally well as the element-wise Huffman.
For the FC1 layer, the reason that vector-sharing comes worse than element-sharing is that the vector size is too small.
Here we select size 2 for all fully connected layers.
The FC1 suffers a poor storage ratio while the other two layers experience enhanced compaction.
Once given a larger vector, as we can see from the Figure~\ref{fig:huffman_vgg16_coarse_vector} that enlarges the size to 4 elements, all FC layers with vector-wise sharing are better than element-wise Huffman.
In general, vector-wise Huffman coding works better than SBSR which is far better than element-wise Huffman coding.

\begin{figure*}
\centering
\begin{subfigure}{0.95\linewidth}
\centering
\includegraphics[width=\textwidth]{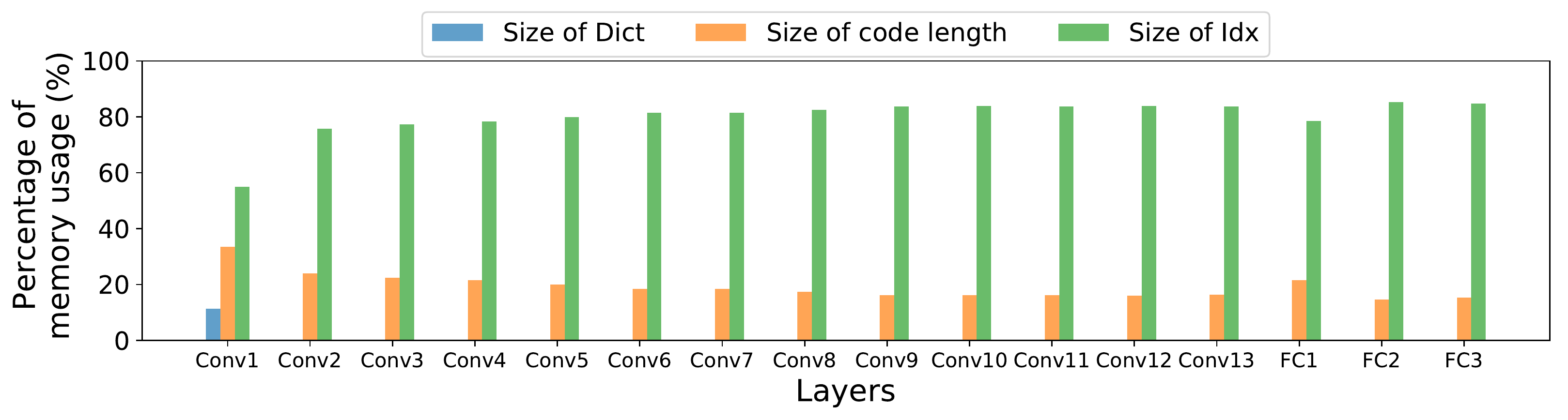}
\caption{Element-wise Huffman Size Breakdown}
\label{fig:huffman_element_breakdown}
\end{subfigure}
\hspace{0.7cm}
\begin{subfigure}{0.95\linewidth}
\centering
\includegraphics[width=\textwidth]{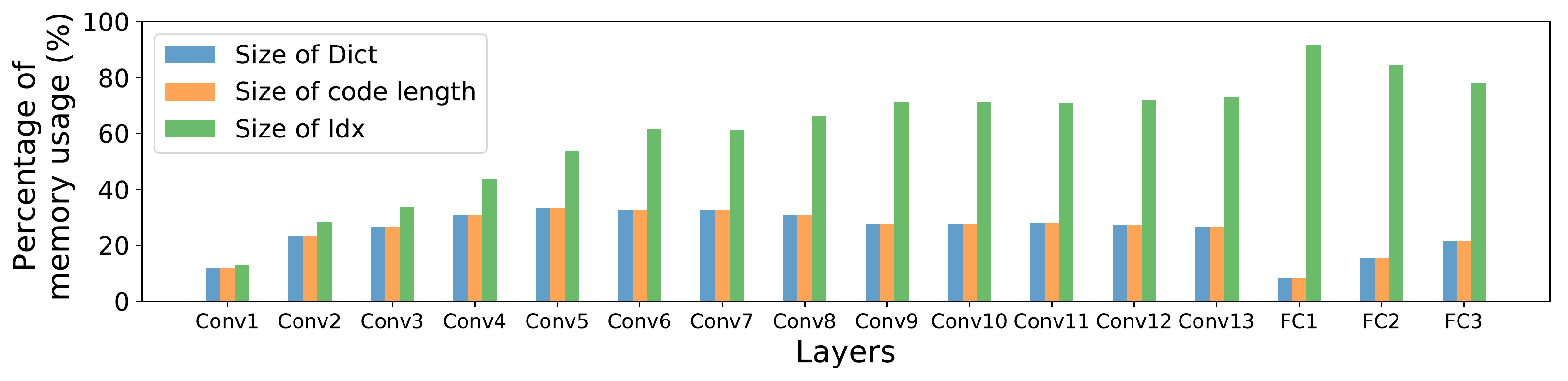}
\caption{Vector-wise Huffman Size Breakdown}
\label{fig:huffman_vector_breakdown}
\end{subfigure}
\begin{subfigure}{0.95\linewidth}
\centering
\includegraphics[width=\textwidth]{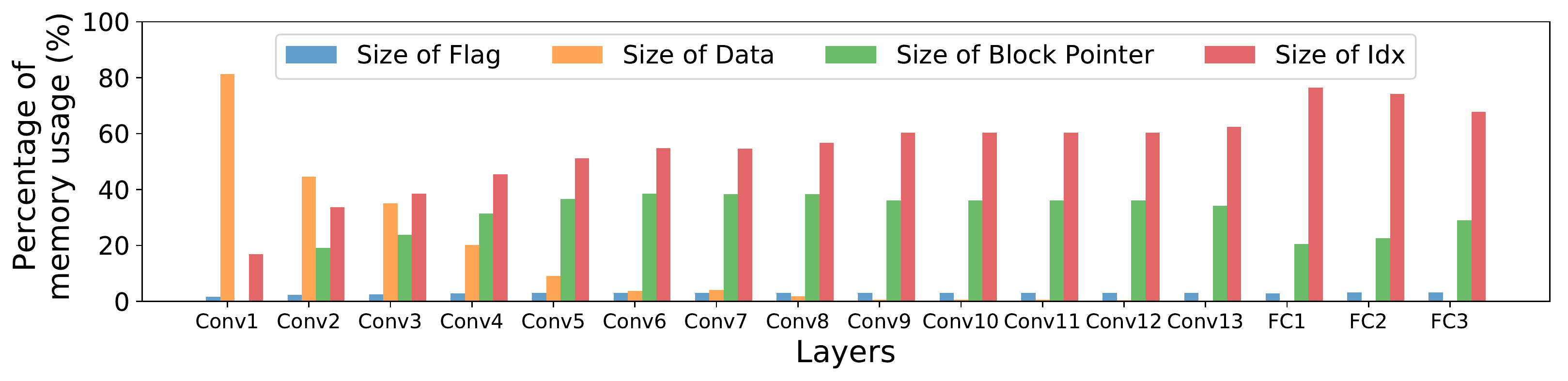}
\caption{SBSR Size Breakdown}
\label{fig:sbsr_breakdown}
\end{subfigure}
\caption{Comparison between the element-wise and vector-wise sharing}
\label{huffman_vgg16_breakdwon}
\end{figure*}

To further understand the memory consumption, we break down the equation 2 and 3 to examine the contribution of each component.
Figure~\ref{huffman_vgg16_breakdwon} shows the percentage of memory required by each item in the equations for element-wise Huffman coding, vector-wise Huffman coding, and the SBSR.
As we can see in the Figure~\ref{huffman_vgg16_breakdwon}, in most of the cases, space spent on storing index dominates the whole memory usage.
However, the memory gap between the index and others are larger for element-wise Huffman Coding, e.g. bars in Figure~\ref{fig:huffman_element_breakdown}, than the vector-sharing methods, e.g. Figure~\ref{fig:huffman_vector_breakdown} and Figure~\ref{fig:sbsr_breakdown}.
Optimising the dominant component can effectively reduce the size of overall memory consumption.

\subsection{Discussion Between SBSR and Huffman coding}
Though both vector-wise Huffman coding and SBSR works better than element-wise sharing, these two methods proposed in this paper have some fundamental difference.
As the Huffman coding builds a binary tree on the vectors according to their repeated frequency, its performance complexity is $O(nlogn)$.
To the contrary, the SBSR does not require to sort the vector. It can be created by a single scan of the tensor; therefore, its computation complexity is $O(n)$.
For extracting the value, both vector-wise Huffman coding and the SBSR has the complexity of $O(1)$.
For Huffman coding, the indices are used to get the code first and then decode it by checking the dictionary.
For the SBSR, the indices are used to get the block in the SBSR.
By checking the flag, we can acquire the value directly or a pointer which leads us to the value.

Apart from the performance complexity, the SBSR are more flexible than the Huffman coding.
For each time when the values changed, we have to rebuild the Huffman tree accordingly.
However, the SBSR format can handle such issue easily.
As it works as a link list, we can insert a new node or simply update the pointer once the vector changed.

In general, there is a compensation between the two implementations we provided in this paper.
The vector-wise Huffman coding provides a better compaction ratio, while the SBSR has a higher performance and flexibility.

%% file: conclusion.tex
\section{Conclusion}

Network pruning and quantization are successful techniques that can
efficiently reduce the size of trained CNN models. However, even after
pruning and quantization there remains significant redundancy in the
form of repeated patterns among the trained parameters. In this paper
we propose a novel approach to compacting trained CNNs by exploiting
this kind of redundancy. We build upon the existing block-sparse row
format for sparse matrices, by sharing a single copy of duplicate
blocks. Repeated blocks are replaced by a reference pointing to their
first appearance. We evaluated our approach on several well-known CNNs
and found that it results in compaction ratios of 1.4$\times$ to
3.1$\times$ in addition to the saving from network pruning and quantization.

We also evaluated element-wise Huffman coding to compress the weight matrices,
and implemented an improved block Huffman coding scheme. Both our SBSR approach
and block Huffman coding improve compression over element-wise Huffman coding
on VGG-16 weight matrices, with an average improvement of $1.53\times$ (SBSR)
and $1.67\times$ (block Huffman coding) across all weight tensors in the
network.